\documentclass[11pt, a4paper]{scrartcl}
\usepackage[margin=2cm]{geometry}
\usepackage[utf8]{inputenc}
\usepackage{booktabs}
\usepackage{subcaption}
\usepackage[skip=2pt, font=footnotesize]{caption}
\usepackage{graphicx}
\graphicspath{{.}{imgs/}}
\usepackage{amsmath}
\usepackage{url}

\title{Designing Behaviour in Bio-inspired Robots Using Associative Topologies of Spiking-Neural-Networks\thanks{Paper submitted to the BICT 2015 Conference in New York City, United States}}
\author{
Cristian Jimenez-Romero\\
 The Open University\\
 MK7 6AA, United Kingdom\\
 cristian.jimenez-romero@open.ac.uk \\
\and
David Sousa-Rodrigues\\
 The Open University\\
 MK7 6AA, United Kingdom\\
 david@davidrodrigues.org \\
\and
Jeffrey H. Johnson\\
 The Open University\\
 MK7 6AA, United Kingdom\\
 jeff.johnson@open.ac.uk
}
\date{}

\begin{document}

\maketitle

\begin{abstract}
 This study explores the design and control of the behaviour of agents and robots using simple circuits of spiking neurons and Spike Timing Dependent Plasticity (STDP) as a mechanism of associative and unsupervised learning. Based on a "reward and punishment" classical conditioning, it is demonstrated that these robots learnt to identify and avoid obstacles as well as to identify and look for rewarding stimuli. Using the simulation and programming environment NetLogo, a software engine for the Integrate and Fire model was developed, which allowed us to monitor in discrete time steps the dynamics of each single neuron, synapse and spike in the proposed neural networks. These spiking neural networks (SNN) served as simple brains for the experimental robots. The Lego Mindstorms robot kit was used for the embodiment of the simulated agents. In this paper the topological building blocks are presented as well as the neural parameters required to reproduce the experiments. This paper summarizes the resulting behaviour as well as the observed dynamics of the neural circuits. The Internet-link to the NetLogo code is included in the annex.
\end{abstract}


\section{Introduction}

With the emergence of third generation artificial neural networks (ANN), better known as "Spiking Neurons", neural networks not only increased their computational capabilities, but also their level of realism with respect to the simulation of biological neurons\,\cite{Gerstner:2002}. 

While most current ANN models are based on simplified brain dynamics, Spiking neurons are capable of simulating a broad (and more realistic) range of learning and spiking dynamics observed in biological neurons such as: Spike timing dependent plasticity (STDP)\,\cite{Gerstner:1996}, long term potentiation, tonic and phasic spike, inter-spike delay (latency), frequency adaptation, resonance, input accommodation\,\cite{Izhikevich2003}. 

In this paper we are especially concerned with one of the characteristics mentioned above, that is: STDP. Our aim is not only to understand how this learning mechanism works at the microscopic level but also how STDP elicit behaviour at a macroscopic level in a predictable way.

A broad body of research has been produced in recent years\,\cite{Gerstner:1996,Bi1998,Song2000}, which describes the dynamics of STDP in populations of Spiking Neurons. 

However, the literature describing the use and implementation of this learning mechanism to control behaviour in robots and agents is not as numerous. 

Circuits of SNNs have been coupled with a double pheromone stigmergy process in a simulation of foraging ants enhancing the behaviour of the simulated swarm.\,\cite{jimenez-romero:2015}.

In work done by\,\cite{Wang2014,Gonzalez-Nalda2011,Helgadottir2013,Cyr2012} circuits of SNN were used to control the navigation of robots in real and virtual environments. STDP and other Hebbian approaches were used as the underlying mechanism of associative learning.

Although in most of the research the spiking dynamics of single and multiple neurons is broadly explained, there is little focus on the topology of the neural circuits. This paper contributes with a model of simple STDP-based topologies of SNN used as building blocks for building controllers of autonomous agents and robots.

\section{Methodology}


A spiking neural network engine was implemented in the multi-agent modelling environment Netlogo\,\cite{tisuenetlogo:2004}. This serves as a platform for building and testing the neural-circuit topologies. The engine is built in the framework of Integrate-and-fire models\,\cite{Gerstner:2002,Izhikevich2003} which recreate to some extent the phenomenological dynamics of neurons while abstracting the biophysical processes behind it. The artificial neuron is modelled as a finite-state machine\,\cite{Upegui2005} where the states transitions (Open and refractory states) depend mainly on a variable representing the membrane potential of the cell. 

The implemented model does not aim to include all the dynamics found in biological models, hence it is not suitable for accurate biological simulations. As there are already robust and scalable tools\,\cite{Bhalla1993,Hines1997,Baxter2007} to simulate large populations of spiking-neurons with complex dynamics. Instead, the model presented here is a SNN engine for fast prototyping of simple neural circuits and for experimentation with small populations of SNN. 


In STDP the synaptic efficacy is adjusted according to the relative timing of the incoming pre-synaptic spikes and the action potential triggered at the post-synaptic neuron: (1) The pre-synaptic spikes that arrive shortly before (within a learning window) the post-synaptic neuron fires reinforce the efficacy of their respective synapses. (2) The pre-synaptic spikes that arrive shortly after the post-synaptic neuron fires reduce the efficacy of their respective synapses.

Eq. \ref{eq:1}\,\cite{Gerstner:1996} describes the weight change of a synapse through the STDP model for pre-synaptic and post-synaptic neurons where: $j$ represents the pre-synaptic neuron, the arrival times of the pre-synaptic spikes are indicated by $t_j^f$ where $f$ represents the number of pre-synaptic spikes $t_i^n$ with $n$ representing the firing times of the post-synaptic neuron:
\begin{equation}
\Delta w_j = \sum_{j=1}^{N}\sum_{n=1}^{N}W(t_i^n-t_j^f)
\label{eq:1}
\end{equation}

The connection weight resulting from the combination of a pre-synaptic spike with a post-synaptic action potential is given by the function $W(\Delta t)$\,\cite{Gerstner:1996,Bi1998,Song2000}
\begin{equation}
W(\Delta t) = \begin{cases}
A_+ \exp({\Delta t}/\tau_+), &\text{if $\Delta t<0$} \\
-A_- \exp({-\Delta t}/\tau_-), &\text{if $\Delta t>0$}
\end{cases}
\end{equation}

where $\Delta t$ is the time interval between the pre-synaptic spike and the post-synaptic action potential. $A_+$ and $A_-$ determine the maximum grow and weaken factor of the synaptic weights respectively. $\tau_+$ and $\tau_-$ determine the reinforce and inhibitory interval or size of the learning window. 


Associative learning is understood as a learning process by which a stimulus is associated with another. In terms of classical conditioning \cite{pavlov1927conditioned}, learning can be described as the association or pairing of a conditioned or neutral stimulus with an unconditioned (innate response) stimulus. 

The pairing of two unrelated stimuli usually occurs by repeatedly presenting the neutral stimulus shortly before the unconditioned stimulus that elicits the innate response. The simplest form of associative learning occurs pair wise between a pre- and a postsynaptic neuron. 
%
%


In order to create a neural circuit of SNNs that allows the association of an innate response to a neutral stimulus, it is necessary to have at least the following elements: (1) A sensory input for the unconditioned stimulus $U$. (2) A sensory input for the conditioned (neutral) stimulus $C$.
(3) The motoneuron (actuator) $M$, which is activated by the unconditioned stimulus.

%

\begin{figure}[ht!]
\centering
\includegraphics[width=8cm]{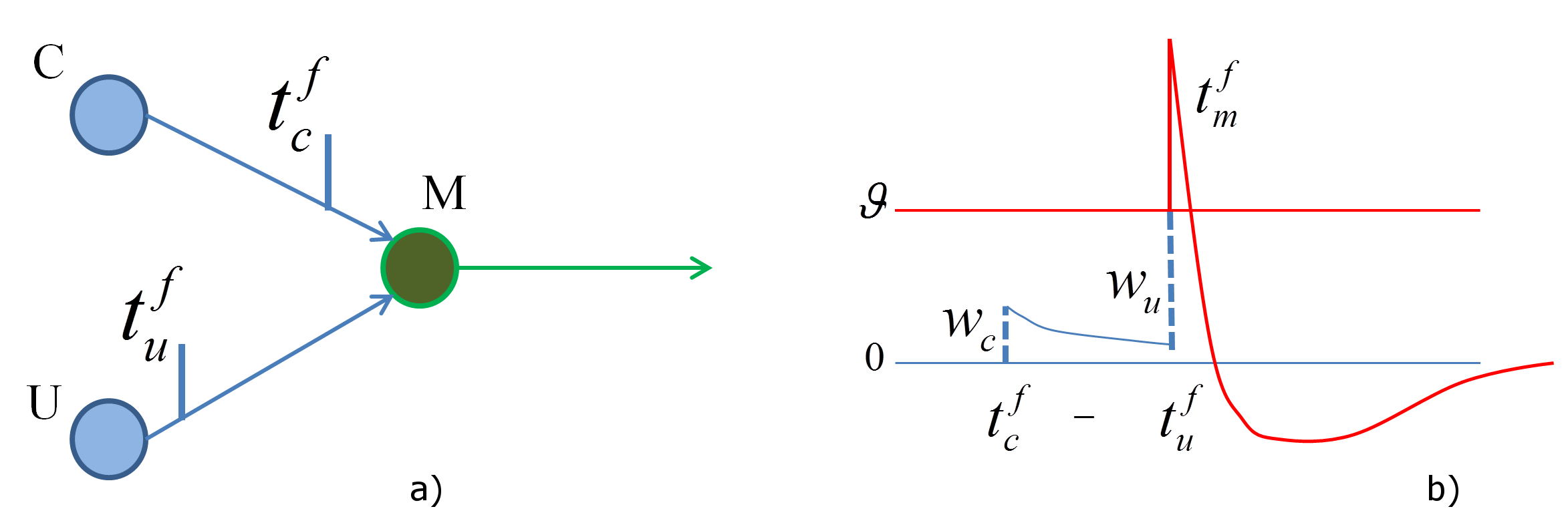}
\caption{a) Excitatory postsynaptic potentials (EPSPs) of amplitude $w_c$ and $w_u$  at times $t_c^f$ and $t_u^f$ respectively. b) Action potential triggered at postsynaptic Motoneuron $M$.}
\label{fig:2}
\end{figure}

The neural circuit in figure \ref{fig:2}a) illustrates the two input neurons $C$ and $U$ each transmitting a pulse to postsynaptic neuron $M$. As shown in \ref{fig:2}b) the unconditioned stimulus transmitted by $U$ triggers an action potential (reaching threshold $\vartheta$) at time $t_m^f$ shortly after the EPSP elicited by $C$ at time $t_c^f$\,\cite{Gerstner:1996,Bi1998,Song2000}.

\begin{figure}[ht!]
\centering
\includegraphics[width=8cm]{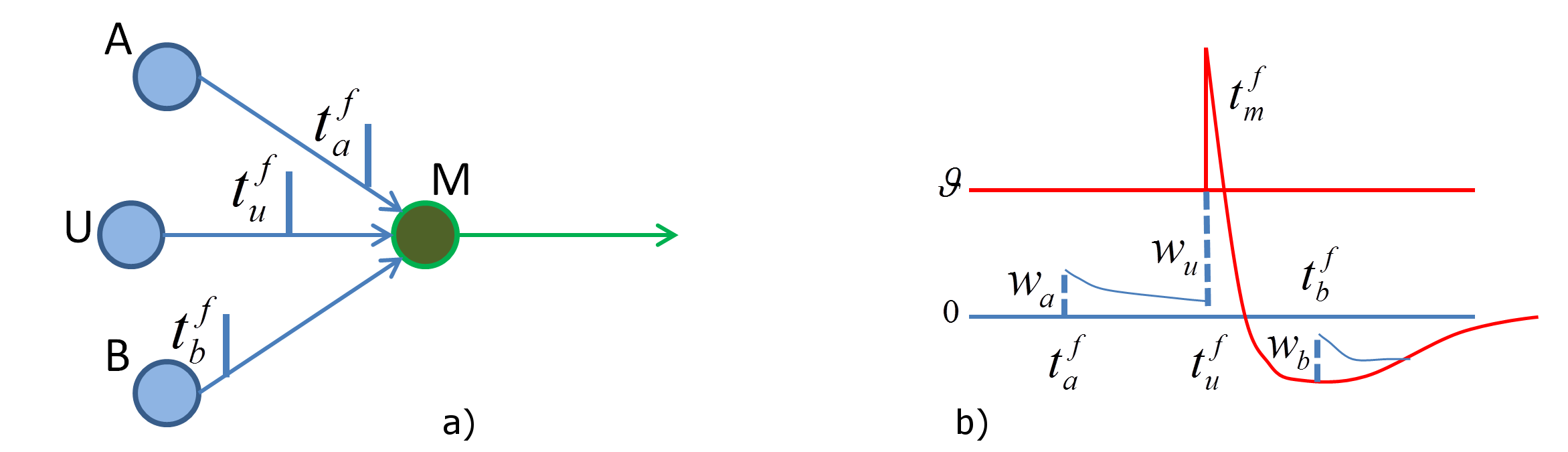}
\caption{a) Spikes emitted by input neurons $A$, $U$ and $B$ reaching the synapse with postsynaptic motoneuron $M$ at time $t_a^f$, $t_u^f$ and $t_b^f$ respectively. b) The spike emitted by $A$ elicits an EPSP of amplitude $w_a$, which is followed a few milliseconds later $(t_u^f-t_a^f)$ by an action potential triggered by $U$ at time $t_u^f$.  The pulse emitted by $B$ arrives shortly after the action potential in $M$ at time $t_b^f$.}
\label{fig:4}
\end{figure}

Given that the STDP learning window allows both LTP and LTD, the simple topology illustrated in figure \ref{fig:2}a), can be extended giving it the ability to associate stimuli from multiple input neurons with an unconditioned response. The topology illustrated in figure \ref{fig:4} includes three input neurons $A$, $B$ and $U$. Neurons $A$ and $B$ receive input from two different neutral stimuli, while $U$ receives input from an unconditioned stimulus.

The circuit in figure \ref{fig:4}a can be used to implement a simple neural circuit to control the movement of an agent or a robot. In such a way that the agent / robot would learn that whenever a (neutral) stimulus in $A$ or $B$ is presented the agent would perform the action associated to $M$. Although, on its own, this circuit only allows a limited margin of actions (trigger reflex or not) in response to input stimuli, this circuit can be taken as a building block which combined in a larger neural topology can produce more sophisticated behaviours.

\begin{figure}[ht!]
\centering
\includegraphics[width=5cm]{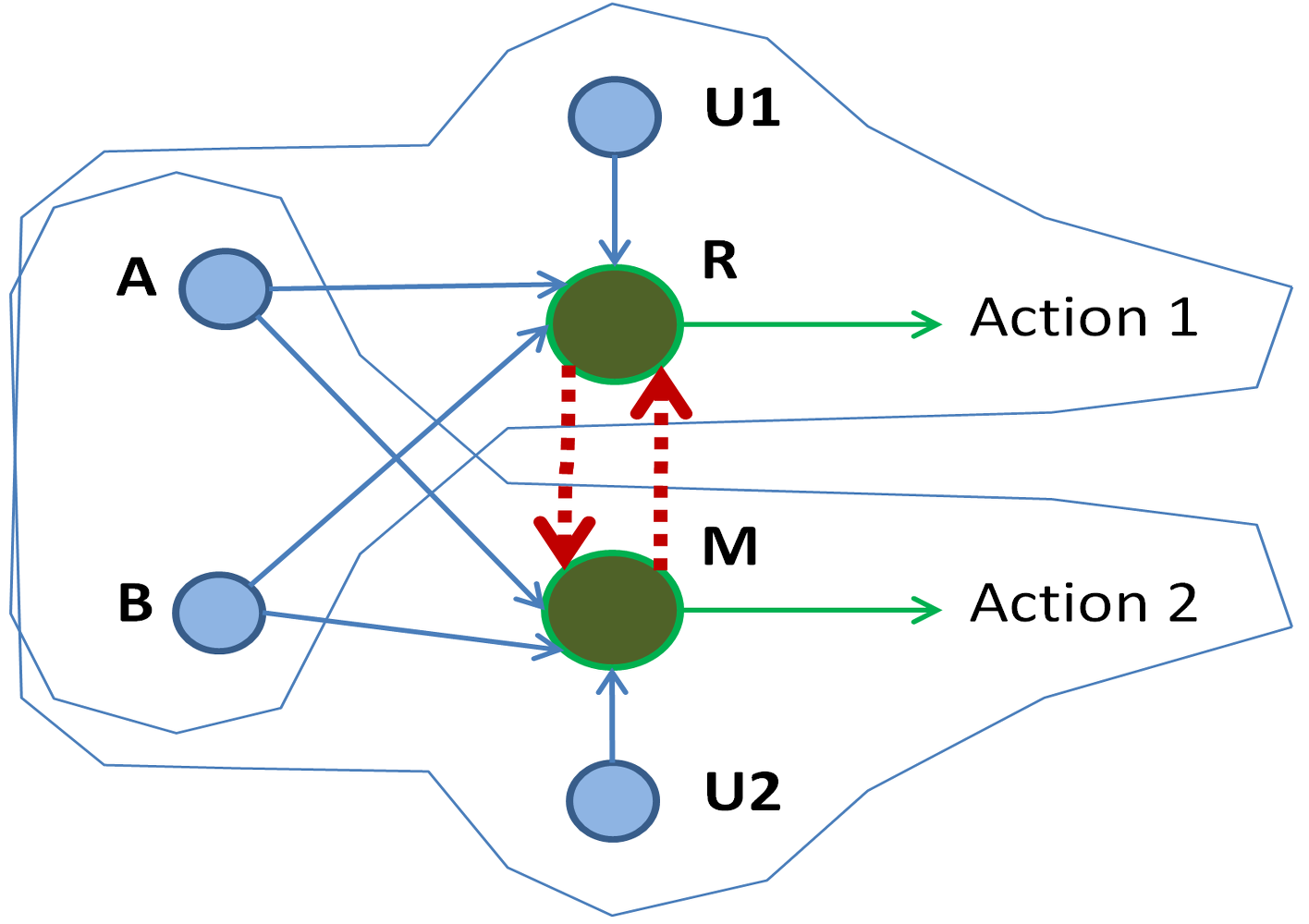}
\caption{Neural circuit with 2 mutually inhibitory sub-circuits.}
\label{fig:6}
\end{figure}

Connecting $A$ and $B$ from the circuit in figure \ref{fig:4} with a second Motoneuron $R$ allows the initially neutral stimuli perceived by neurons $A$ and $B$, to be associated to the corresponding actions elicited by $R$ and $M$. The new neural circuit with 2 motoneurons is illustrated in figure \ref{fig:6}. 


The top part contains the sub-circuit which creates the association between the input stimuli received in $A$, $B$ and the action elicited by $R$ (Action 1). While The bottom part contains the sub-circuit which creates the association between $A$, $B$ and the action elicited by $M$ (Action 2). Although both sub-circuits share the same input neurons $A$ and $B$, the elicited behaviour in $R$ and $M$ will depend on the firing-times correlation between the neutral (conditioned) inputs $A$, $B$ and the unconditioned neurons $U1$ and $U2$.

In figure \ref{fig:6} both Actions 1 and 2 can be performed at the same time if the same inputs in the top and bottom parts are reinforced in both sub-circuits. This behaviour however can be inconvenient if the system is expected to perform one action at the time. Inhibitory synapses between sub-circuits provide a control mechanism in cases where actions are mutually exclusive. For this, the mutually inhibitory synapses in Motoneurons $R$ an $M$ work as a winner-take-all mechanism where the first firing neuron elicits its corresponding action while avoiding the concurrent activation of other sub-circuit(s).  


The neural circuit in figure \ref{fig:6} was used as a model to implement in Netlogo a simple micro-brain to control a virtual insect in a simulated two dimensional environment. The simulated micro-brain was able to process three types of sensorial information: (1) olfactory, (2) pain and (3) pleasant or rewarding sensation. The olfactory information was acquired through three receptors where each receptor was sensitive to one specific smell represented with a different color (black, red or green). Each olfactory receptor was connected with one afferent neuron which propagated the input pulses towards the Motoneurons. Pain was perceived by a nociceptor whenever the insect collided with a wall (black patches) or a predator (red patches). Finally, a rewarding or pleasant sensation was elicited when the insect came in direct contact with a food source (green patches).

\begin{figure}[ht!]
\centering
\includegraphics[width=8cm]{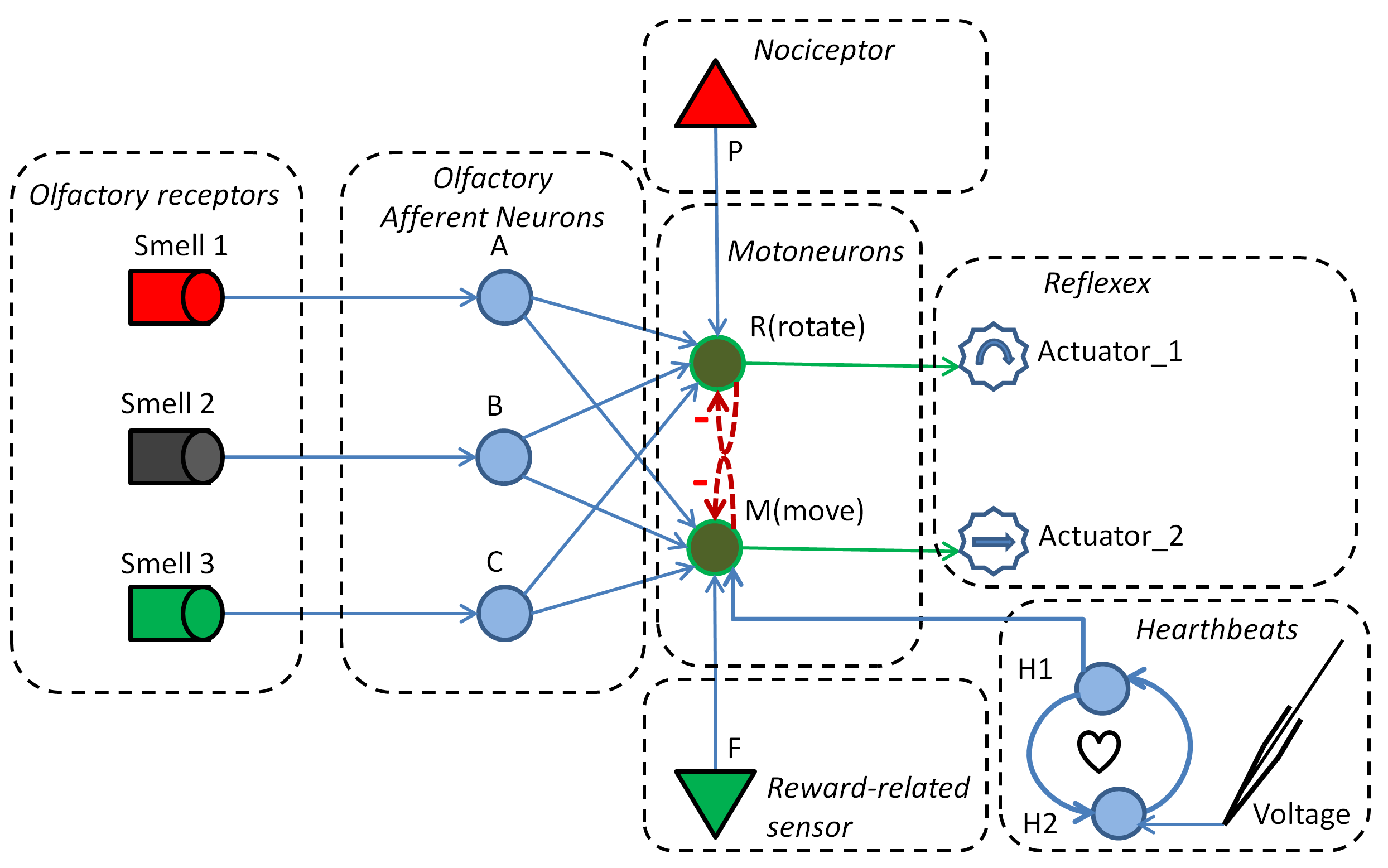}
\caption{Neuro-inspired controller of the virtual insect}
\label{fig:7}
\end{figure}

The motor system is equipped with two types of reflexes: 1) Rotation and 2) Moving forward. Both actions produced by Actuator\_1 and Actuator\_2 respectively. The number of rotation degrees as well as the number of movement units were set in the simulation to $5^{\circ}$ and 1 patch respectively. In order to keep the insect moving even in the absence of external stimuli, the motoneuron $M$ was connected to a sub-circuit composed of two neurons $H1$ and $H2$ performing the function of a pacemaker sending periodic pulses to $M$. Figure \ref{fig:7} illustrates the complete neural anatomy of the virtual insect.

%

The simulation environment was connected with a Lego Mindstorms EV3 robotic platform\,\cite{mindstorms2015} with the the following architecture, which served as embodiment for the simulated virtual insect described above:
The olfactory system of the insect was simulated using the EV3 colour sensor camera positioned in front of the robot and looking towards the floor. If the captured colour was black, red or green, the respective receptor in the neural circuit was activated. The nociceptive input was simulated using the EV3 ultrasonic sensor  positioned in front of the robot. This sensor reported distance to objects and values less than 5 cm were assumed to be as collision and  consequently the nociceptor in the neural circuit was activated. The reward input was simulated using the EV3 touch sensor positioned on top of the robot. In case of pressing, it activated the reward receptor of the neural circuit. The movement and rotation of the robot was controlled by a differential drive assembly of the motors. When the motoneuron $M$ fired, the simulation environment sent a forward command to the EV3 platform for 500 milliseconds. When the motoneuron $R$ fired, the simulation sent a command to the EV3 platform requesting the activation of both servo motors rotating in opposite directions, resulting in a spin of the robot. 
The floor was made up of coloured squares including the three associated to the nociceptive and rewarding stimuli. Other colours were taken by the robot as empty space. Objects of the same hight as the ultrasonic sensor were positioned in the centre of the black and red squares. This aimed to activate the nociceptor of the neural circuit every time the robot came closer to the black and red patches. The touch sensor was manually pressed by the experimenter every time the robot moved over a green square. This activated the reward receptor of the neural circuit.

\section{Results}

\begin{figure}[ht!]
\centering
\begin{subfigure}[b]{0.22\textwidth}
\includegraphics[width=\textwidth]{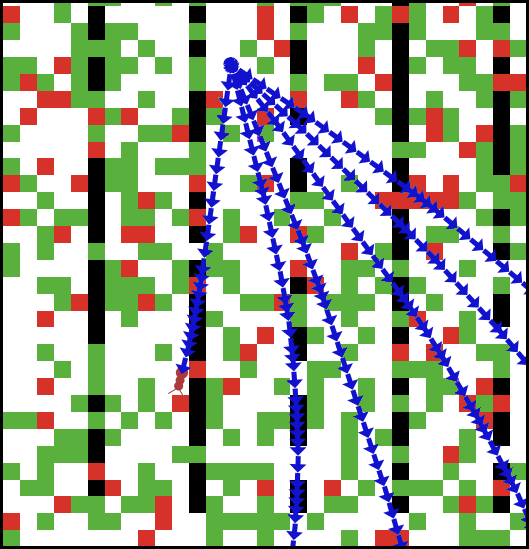}
\label{fig:8}
\end{subfigure}
\begin{subfigure}[b]{0.22\textwidth}
\includegraphics[width=\textwidth]{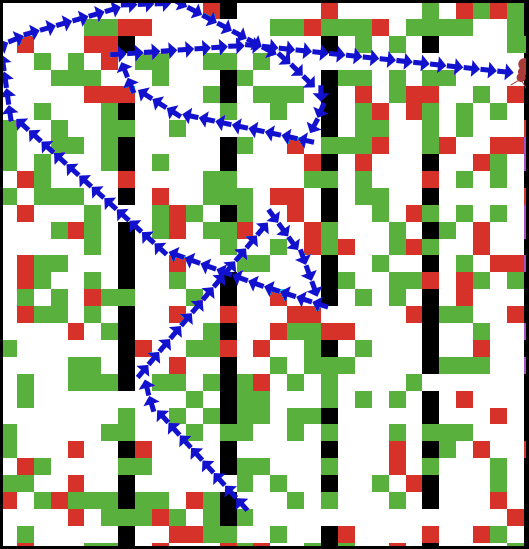}
\label{fig:9}
\end{subfigure}
\caption{Short trajectories at the training phase. Insect collides and escapes the world repeatedly (left). Long trajectory shows insect avoiding red and black patches (right).}
\label{fig:89}
\end{figure}

At the beginning of the training phase (Figure \ref{fig:89} left) the insect moves along the virtual-world colliding indiscriminately with all types of patches. The insect is repositioned on its initial coordinates every time it reaches the virtual-world boundaries. As the training phase progresses it can be seen that the trajectories lengthen as the insect learns to associate the red and white patches with harmful stimuli and consequently to avoid them (See Figure 8 right). After approximately 15000 iterations, the insect moves collision free completely avoiding red and black patches while looking for food (green patches). 

The artificial insect is able to move collision free after about 15 thousand simulation iterations. This number depends on the parameters set for the circuit neural-dynamics and the STDP learning rule. Table 1 shows the learning behaviour in terms of iterations required for a collision free movement, using different values for the learning constants A+ and A- (eq. 3) to respectively potentiate or depress the synaptic weights between the afferent and Motoneurons:

\begin{table}[ht!]
\centering
{\scriptsize
\caption{Learning behaviour with different $A_+$ and $A_-$ parameters}
\label{table:1}
\begin{tabular}{@{}cc@{}}
\toprule
\begin{tabular}[c]{@{}c@{}}Symmetric LTP/LTD\\ A+, A-\end{tabular} & \begin{tabular}[c]{@{}c@{}}Number of iterations before collision\\ free movement\end{tabular} \\ \midrule
0.01                                                               & 19,000                                              \\
0.02                                                               & 15,000                                              \\
0.03                                                               & 9,000                                               \\
0.04                                                               & 7,000                                               \\ \bottomrule
\end{tabular}
}
\end{table}

The behaviour observed in the simulation was reproduced with the EV3 robot. However, it was necessary to adjust the parameters A+, A- to 0.08 and the number of rotation and movement units in order to speed up the training phase given that in the simulation environment the neural circuit iterates at about 2000 steps per second while in the real world the robot was interacting with the neural circuit at about 50 iterations per second. The lower iteration speed was was an optimisation issue in the communication interface between the robotic platform and the simulation environment which was programmed by the experimenters. In any case the robot was able to show the same learning and adaptation abilities originally observed in the simulated insect. 

\section{Conclusion}

SNN mimic their biological counterparts in several ways but possibly their most relevant characteristic is their ability to use spatio-temporal information for communication and computational purposes in a similar way to biological neurons. 
With their ability to represent information using both rate and pulse codes, SNN become an efficient and versatile tool to solve several time dependent problems. 
Although some traditional ANNs can include temporal dynamics by the explicit use of recurrent connections, the inherent notion of time in the nature of SNNs makes them by design closer to the biological observed entities. This makes the dynamics of SNN more plausible than those of traditional ANNs. 
SNNs are becoming very efficient because they are capable of replacing large ensembles of traditional ANNs. This makes them very suitable for application in situations where high performance and low power 
consumption are important. In robotics this is of particular interest as reducing power consumption and increasing computational power mean higher levels of autonomy and performance in situations where robots are operating in real time or near to real time. 
The impact of SNNs new computational model will be key in the development of new bio-inspired robotics and new artificial agents, allowing for unprecedented evolution in the field. The model presented in this paper is a first step in showing how to design and control the behaviour of agents and robots using simple circuits of spiking neurons, and it will hopefully seed future developments in the area.

\section*{Software}

The Netlogo simulation software SpikingLab is available at \url{http://modelingcommons.org/browse/one_model/4455}

\bibliographystyle{abbrv}
\bibliography{references} 

\end{document}